# Collapse Resistant Deep Convolutional GAN for Multi-Object Image Generation

Elijah D. Bolluyt
Electrical and Computer Engineering
Stevens Institute of Technology
Hoboken, NJ, USA
ebolluyt@stevens.edu

Cristina Comaniciu
Electrical and Computer Engineering
Stevens Institute of Technology
Hoboken, NJ, USA
ccomanic@stevens.edu

*Abstract*—This work introduces a novel system for the generation of images that contain multiple classes of objects. Recent work in Generative Adversarial Networks have produced high quality images, but many focus on generating images of a single object or set of objects. Our system addresses the task of image generation conditioned on a list of desired classes to be included in a single image. This enables our system to generate images with any given combination of objects, all composed into a visually realistic natural image. The system learns the interrelationships of all classes represented in a dataset, and can generate diverse samples including a set of these classes. It displays the ability to arrange these objects together, accounting for occlusions and inter-object spatial relations that characterize complex natural images. To accomplish this, we introduce a novel architecture based on Conditional Deep Convolutional GANs that is stabilized against collapse relative to both mode and condition. The system learns to rectify mode collapse during training, self-correcting to avoid suboptimal generation modes.

## I. Introduction

Generative Adversarial Networks (GANs) [1] have succeeded in a wide array of image generation tasks. Many of these tasks require Conditional GANs [2], a specialized type of GAN that conditions each generated sample on an input that specifies some properties or contents the sample should contain. Conditional GANs can use a single model to generate multiple classes of samples, even specifying multiple characteristics that the output image should possess, as with the generation of human faces [8]; these models can be directed to produce a highly specific output based on the condition, such as translating from one image class to another while keeping the image's structure intact [6]. They can also translate between domains, using complex conditional inputs such as text [5] and images [4] processed by a deep architecture, and can generate a corresponding sample in another domain. Many GAN systems generate images that each have a single object as the focus of the generative task, such as the human face or individual animals. Those that generate multi object images usually use a complex conditional input to provide the details of the visual structure, such as the translation of semantic segmentation maps to natural images [4] or image to image translation networks [6] where the conditional input provides significant information about the visual structure of the output. Many of these works use some variant of the Deep Convolutional GAN (DC-GAN) architecture [3], comprised of fully convolutional deep neural networks serving as both the Generator and the Discriminator.

This work introduces a novel Conditional GAN architecture that is capable of generating images containing multiple classes of objects, composed in a visually correct structure, using only a list of class labels as the conditional input. This novel task extends the current Conditional GANs beyond generating multiple classes using a single model, also learning to compose images that mesh multiple classes together; the system must learn not only how to generate realistic images of each class, but also their structural interrelationships in the visual space.

In previous systems, this information is usually provided by an image or other sample given as a conditional input. The lack of this information increases the complexity of the task at hand by requiring the system to learn how to properly structure a natural image, and how to insert multiple classes into it in a realistic composition. Omitting this information extends the system's capabilities to generation of multi-class images without being given information on the image's desired structure; this is desirable because most domains of unprocessed natural images contain many classes of objects, so this task must be addressed for GANs to succeed in generating more complex natural images without the need for a supervising complex conditional input.

Training GAN systems poses a few unique challenges, including problems achieving convergence, and mode collapse, where the Generator sacrifices diversity in its output samples in order to overcome the Discriminator without learning the full complexity of the desired output space. The issue of collapse can become even more pernicious in conditional generation; in mode collapse, the Generator mitigates the effect of the random noise input on its output, which leads to a homogenous output set. In conditional generation, the Generator may also learn to ignore the conditional input, resulting in output samples that do not closely correspond to the desired properties or classes specified in the condition.

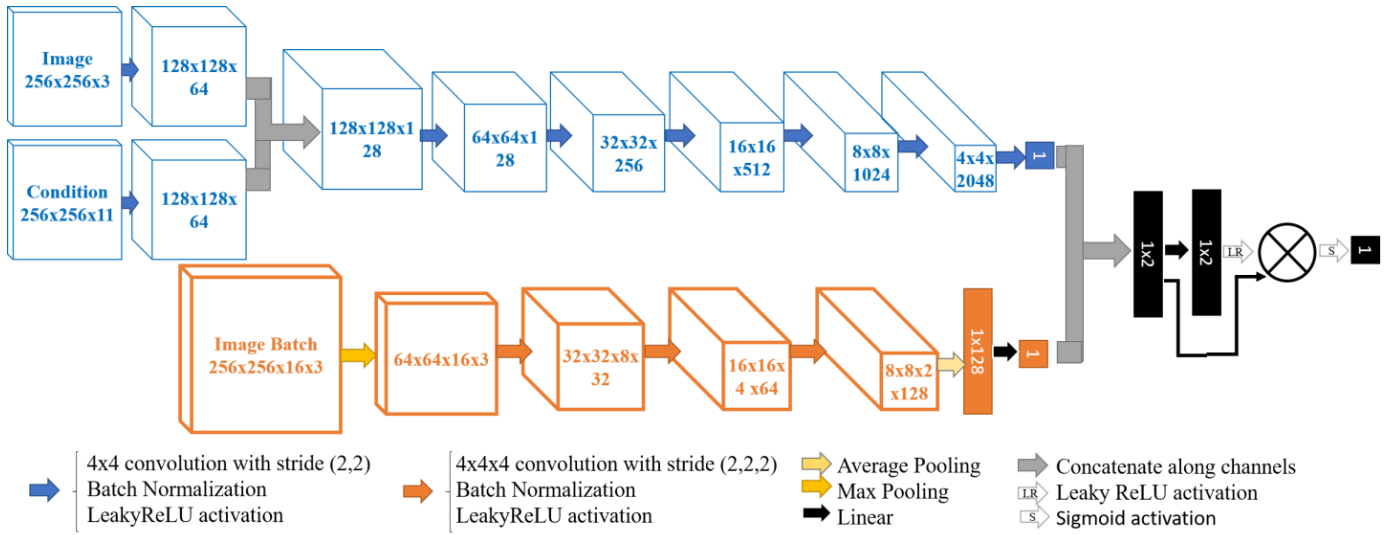

Figure 1. Modified Discriminator System Diagram

Multi-class generation compounds this problem further by making the samples more homogenous. A set of images, each containing several of a small number of classes, significantly overlap one another in contents, making each class more difficult for the system to distinguish. This homogeneity among images containing different classes significantly increases the complexity of generating realistic images. In addition, objects in natural images usually overlap and occlude each other, making the individual classes unrecognizable by shape alone, requiring the system to learn to produce and recognize objects based partially on shape but often mostly on texture and visual features.

These difficulties intensify the problems already present in training a GAN to converge properly, and render existing systems for single class generation unsuitable to this task. Our system overcomes these difficulties by introducing a novel architectural feature to the Discriminator, while maintaining the Generator architecture from the Conditional DCGAN. Our solution is based on the concept of minibatch discrimination [9], or sharing information between samples within generated batch in order to mitigate some forms of training collapse. Our novel method of implementing this concept involves adding to the Discriminator a parallel branch to the main DCGAN branch which samples each batch output by the Generator as a single three-dimensional volume. This structure uses three dimensional convolutions to evaluate each batch as a whole, providing insight to the system on the distribution the Generator is outputting, rather than its individual samples.

This gives the Discriminator the ability to easily distinguish relatively homogenous batches, and thus steer the system away from collapse. The system is equipped with an attention mechanism to weight the importance of the two branches dynamically; this allows the Discriminator to emphasize the three-dimensional branch to pull the Generator out of a collapse, and to deemphasize it during normal training.

## II. MODEL

### A. DC-GAN

The Deep Convolutional GAN architecture [3] uses fully convolutional deep neural networks for both the Generator and the Discriminator. The Generator utilizes fractionally strided convolution operators to upsample a noise vector to an image, and the main branch of the Discriminator is composed of strided convolution operators to downsample the images to a decision vector. Both are fully convolutional, containing no pooling layers. They use the Leaky ReLU non-linear activation function and Batch Normalization.

### B. Conditional DC-GAN

DC-GAN can be extended to conditional generation by allowing for the input of a condition vector to both the Generator and the Discriminator. In both cases, the input vector is reshaped to the same dimensionality as the primary input to that network; for the Generator, the condition is given as a vector to be upsampled in the same manner as the noise vector. For the Discriminator, the conditional vector is extended to the same spatial resolution as the images. In both cases, the condition is fed into a separate first layer, and the output is concatenated with the output of the first layer in the main branch of the architecture and fed into the second layer together. From here, the conditional system functions identically to the non-conditional original.

### C. Extending DC-GAN Generator for multi-class images

In Conditional DC-GAN, a one-hot vector class label is passed to the Generator and the Discriminator; this vector serves as the condition on the generated sample. For multi-class generation, this condition is extended to be a summation of the one-hot vectors for each class represented in the current sample. This maintains the condition as a binary vector of fixed length, compatible with the original formulation, while allowing the arbitrary specification of any combination of classes for generation. This construction also gives the system the minimum amount of information possible, stating only the desired classes and nothing about their relationships or structure. This requires the system to learn the classes' interdependent placement and characteristic occlusion behavior, common in natural images.

TABLE I. CLASS GROUPS (CITYSCAPES DATASET)

| Class | | | | |
|---|---|---|---|---|
| 0 | 1 | 2 | 3 | 4 |
| Road | Ground | Buildings | Fences/ Railings | Traffic lights, light poles |
| 5 | 6 | 7 | 8 | 9 | 10 |
| Nature | Sky | People | 4w Vehicles | Bikes | Trains |

### D. Modified Discriminator Architecture

The Discriminator contains two branches, one taken from the DC-GAN architecture, and a novel branch that helps to avoid collapse (Fig. 1). This new branch performs minibatch discrimination, using information from all samples in a batch to inform the Discriminator's evaluation of each individual sample. This allows the Discriminator to stabilize its evaluations and distinguish real minibatches from fake ones when the Generator enters a collapsed failure state.

This second branch is fully convolutional, using strided three-dimensional convolution operators with Leaky ReLU activation and Batch Normalization. It concatenates each batch of images into a single three dimensional volume and operating on it with three dimensional operations. This allows information from all samples to inform the Discriminator's decision on each sample's veracity. By evaluating the batch as a unit, this structure has the ability to learn to detect both types of collapse, helping to escape non-convergent states, and backpropagate appropriate corrective measures to the Generator. Each layer of this branch has many fewer channels, and the branch has far fewer overall parameters, then the main DCGAN branch, as it is intended only for use to avoid collapse, not to provide meaningful visual feature correction to the Generator. The main DC-GAN architecture should still provide most of the significant feedback to the Generator.

The Discriminator's output is a weighted sum of the two parallel branches' outputs.

$$D = \alpha D_1 + \beta D_2 \quad (1)$$

Since the three-dimensional branch has less ability to provide helpful feedback during non-collapsing training due to its less robust parameter count, the Discriminator should learn to weight this branch more heavily when the Generator trends towards collapse, and less heavily when the system is training properly. This makes a fixed weighting unsuitable, so the system dynamically weights the two branches via an attention mechanism. A single fully connected layer with Leaky ReLU activation takes the output of each branch, a single unnormalized value, and generates a weight for each branch's output. The weighted sum of these two outputs is the final output decision of the Discriminator.

$$[\alpha\ \beta] = \text{LeakyReLU}\left(\begin{bmatrix} a_{11} & a_{12} \\ a_{21} & a_{22} \end{bmatrix} \begin{bmatrix} D_1 \\ D_2 \end{bmatrix} + [b_1\ b_2]\right) \quad (2)$$

Where $a_{ij}$ and $b_i$ are learnable parameters of the system.

This allows the Discriminator to have full control over how the two branches are weighted, and to evaluate during training how close to collapse the system is, and thus how heavily to weight the three-dimensional branch.

The system was trained using the Adam optimizer [12], using the standard GAN criterion [1], which minimizes the JS-Divergence between the real distribution and the generated distribution.

### E. Comparison to existing methods

Many solutions have been proposed to alleviate problems of collapse and instability in GANs, but many, such as Wasserstein GAN with Gradient Penalty (WGAN-GP) [10] introduce some form of normalization to the system in the form of an additional loss term; in our experiments, we found that these penalty functions often slowed convergence of the system and increased the difficulty of the generation task. Our system adds complexity to the system in the form of additional learnable parameters but does not impose any additional losses to be optimized against, maintaining the original GAN target function. This allows the system to be unconstrained by any objective other than the competition between Generator and Discriminator. It does, however, implicitly modify the objective to penalize low entropy sample distributions, a desirable property to avoid collapse.

## III. EXPERIMENTS

### A. Baseline DC-GAN Architecture on multi-class MNIST

We conducted several series of experiments to analyze the task of multi class generation using GANs. We began by training a vanilla DCGAN architecture on the conditional generation of pairs of digits from the MNIST dataset [11]. We prepared the data by embedding two randomly selected samples into a single grayscale image; the digits were placed side by side, with no overlap, at their original resolution. The sum of the two one-hot vectors representing the value of each digit in the image was used as the conditional input. This system converged quickly, generating high quality samples that accurately reflected the desired digits (Fig. 1).

### B. Baseline DC-GAN Architecture on Cityscapes Dataset

The next experiment used the Cityscapes dataset [7] with the same vanilla DCGAN architecture and the same type of conditional input, a sum of one-hot vectors denoting the classes present in the image. For this experiment, we condensed the classes present in the dataset to ten groups, given in (Table 1); we downsampled the images to a resolution of 256x256px due to memory constraints.

This system failed to produce meaningful images, despite many attempts to balance the Generator and Discriminator networks by varying several hyperparameters, including the number of learnable parameters, the learning rate of each network, and the relative number of iterations per network. In every attempt to train the system, it would collapse into a state in which it generated images that were primarily noise, with a few ill-defined shapes present. In these states, the generated batches contained little to no variation among the samples, illustrating a form of collapse that ignores both the random and conditional inputs to the Generator as well as a lack of visual clarity. The system displayed loss values that appeared convergent at these states, with the Discriminator unable to accurately distinguish the real samples from generated ones, and unable to provide meaningful feedback to the Generator network.

This pair of experiments illustrates the difficulty in generating multi-class images with high visual complexity using the baseline DCGAN architecture. While the baseline system could generate images containing easily separable, highly differentiated classes, as in our experiment with multi class MNIST samples, it was unable to learn to distinguish and produce the individual classes that comprise a complex natural image.

### C. Modified DC-GAN on Cityscapes Dataset

We then implemented the proposed novel system, utilizing the additional branch and attention mechanism in the Discriminator network while keeping the baseline DCGAN Generator architecture. We trained this system again on the Cityscapes dataset, with the same conditional input vectors as the previous experiment. We used the Adam optimizer [12] with a learning rate of 0.0002 and $\beta_1 = 0.5$, $\beta_2 = 0.999$.

The system was able to learn to produce visually coherent images that accurately reflected most of the classes specified in the conditional input vector [Figure 3]. The system was unable to learn three out of eleven of the classes; at the downsampled resolution required to process the images, people and bicycles (classes 7,9) were too small for the system to accurately recognize and produce, and trains (class 10) were not present in enough samples for the system to learn to properly distinguish them from other classes. For all other combinations of classes, the system was able to produce visually compelling results, composing the various classes properly into a natural image. It shows the ability to not only place the objects in correct locations, but also to blend them together into a natural scene, complete with sensible occlusions expected in a layered natural image. The system showed resiliency to both mode collapse and collapse relative to the conditional input, producing varied results that closely adhered to the given condition vectors. We generated and evaluated 150 samples, all of which conformed closely to the conditional input for the eight classes that were learned successfully. They displayed an estimated 90% diversity, with approximately 10% repeated images. Figure 4 contains 36 generated samples from this set, illustrating their wide variation.

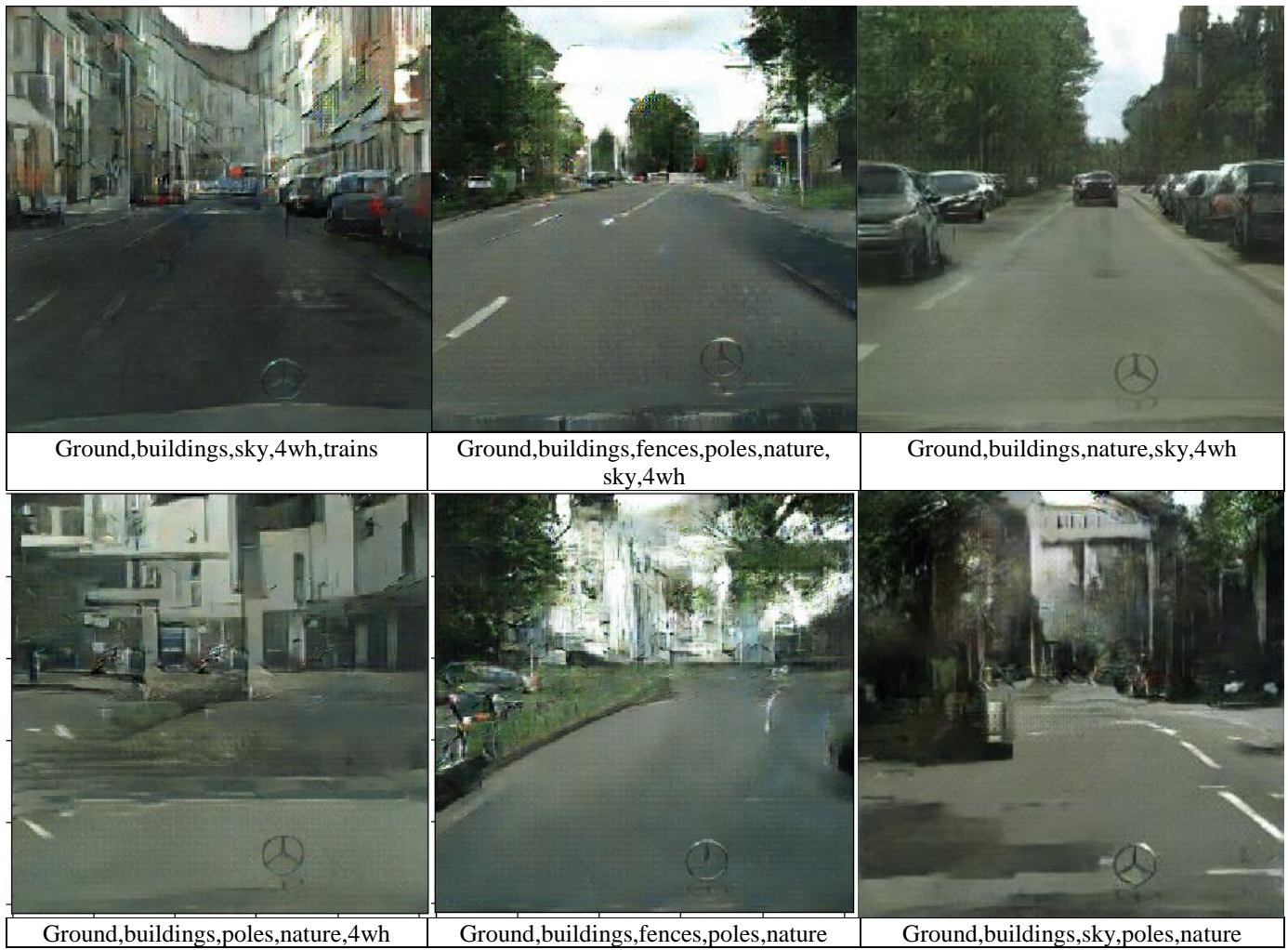

| Ground,buildings,sky,4wh,trains | Ground,buildings,fences,poles,nature,sky,4wh | Ground,buildings,nature,sky,4wh |
| Ground,buildings,poles,nature,4wh | Ground,buildings,fences,poles,nature | Ground,buildings,sky,poles,nature |

Figure 3: Generated Samples from Modified DC-GAN

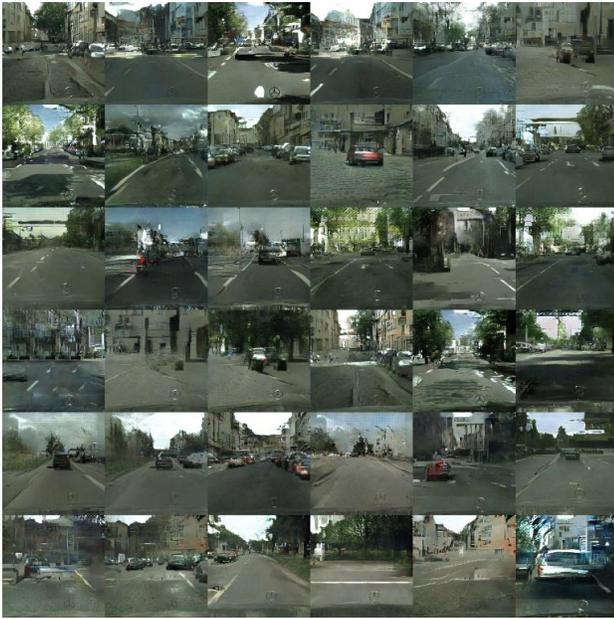

Figure 4: 36 Generated Samples from Modified DC-GAN

## IV. Conclusion

This work introduces a novel architectural structure to enable the training of GAN systems on tasks of multi-class generation, without the need for a complex conditional input providing information on the structure of the output image. The proposed system was able to compose multiple classes of objects into a single image, given no information on desired structural features beyond inferences from the training dataset. Our novel discriminator architecture utilizes minibatch discrimination, sharing information among samples within a batch to stabilize its training, allowing the system to avoid mode collapse as well as collapse relative to the conditional input.